# Clustering based rewarding algorithm to detect adversaries in federated machine learning based IoT environment

Krishna Yadav, B.B Gupta, Senior *Member, IEEE*
Affiliation: National Institute of Technology Kurukshetra
krishna.nitkkr1@gmail.com, gupta.brij@gmail.com

*Abstract--* In recent times, federated machine learning has been very useful in building the intelligent intrusion detection system for IoT devices. As IoT devices are equipped with a security architecture vulnerable to various attacks, these security loopholes may bring a risk during federated training of decentralized IoT devices. Adversaries can take control over these IoT devices and inject false gradients to degrade the global model performance. In this paper, we have proposed an approach that detects the adversaries with the help of a clustering algorithm. After clustering, it further rewards the clients for detecting honest and malicious clients. Our proposed gradient filtration approach does not require any processing power from the client-side and does not use excessive bandwidth, making it very much feasible for IoT devices. Further, our approach has been very successful in boosting the global model accuracy, up to 99% even in the presence of 40% adversaries.

## I. INTRODUCTION

In recent decades, people have seen rapid growth in IoT devices. The requirement of less power and computational infrastructure, remote control access, and low price have made IoT devices popular in health informatics, transportation, home automation, and other sectors [1-2]. Due to the limitation of computational infrastructure, today's IoT devices are not embedded with several layers of security architecture [3]. This weak security architecture has been a central target of adversaries to malfunction IoT devices. With the rise of novel attacks from adversaries, traditional blacklisting techniques for intrusion detection in IoT devices is no more reliable. In recent times, extensive amounts of work can be found in the literature to overcome this problem, which uses different machine and deep learning algorithms for intrusion detections [4-5].

Machine and deep learning-based intrusion detection technique requires large amounts of data to be centrally aggregated and processed [6]. Intrusion detection based dataset are mostly the logs of network and web. According to [7], recent data breach activities have led to the exposure of this data to the adversaries. Adversaries use these logs to find out some useful information to launch even some more devastating attacks against users [8].

Considering the privacy of people, Google, in 2017, has introduced federated machine learning [9]. Federated machine learning trains the global model with the dataset present at the client's side. Here, every client in federated training is decentralized around the world. The federated environment has protected people's privacy by not exporting the client's dataset at the server side. Moreover, federated learning has also helped to build better machine learning models by discovering the knowledge from the data produced by billions of devices around the world.

As the IoT nodes are decentralized, the central authority for injecting data and gradients remains in the hand of these IoT nodes. As IoT devices are much vulnerable to the attacks [10], adversaries can take advantage of security loopholes in these IoT devices to inject false data and gradients. Adversaries may inject these false gradients at any time, i.e., at the client-side, during communication between client and server, and at the server-side. An attack is possible on the client-side by gaining control over the clients, attack during communication round is possible by adversary acting as a man in the middle during federated training, and at the server side is possible by gaining control over the server. The wrong gradients result in the inference of false knowledge from the dataset generated from IoT devices. These false gradients may require higher communication rounds for a global model to converge during a federated environment. A higher communication round will cost substantial computational power and bandwidth, which is beyond the capability of resource constrained IoT devices. Sometimes highly noised gradients may lead to a significant decrease in model performance. A sophisticated gradient poisoning attack in machine learning-based IDS is a source of several other attacks in billions of IoT devices, such as DDOS attacks and phishing attacks. These attacks are possible because gradient poisoning directly affects the capacity of detecting intrusion in deployed IDS in IoT devices. At this time, it becomes very essential to develop an efficient approach which can deal with these kinds of gradient poisoning attacks at the same time feasible for resource constrained IoT devices.

This paper has introduced a clustering-based algorithm that filters the updated gradients on the server-side. The gradient filtration is done by rewarding every client's positive and negative score based on honest and malicious ID obtained from the K-means clustering algorithm. Reward-based algorithms further helps to identify the adversaries that hinder the global models performance at a large scale and remove them permanently. A reward-based algorithm will decrease the communication round, and the accuracy of a global model will also be increased.

The rest of the paper is organized as follows: Section II gives a literature overview of efforts that have been made to solve gradient poisoning attacks in a federated environment. Section III gives background knowledge on federated learning and averaging. Section IV briefly discusses our proposed clustering-based algorithm. Section V provides an overview of our experimentation methodology, whereas section VI briefly discusses our results. Finally, section VII concludes the paper.

---

¹If there is a sponsor acknowledgment, it goes here.

## II. RELATED WORK

Federated machine learning is itself a very new concept in the literature. Very little research work can be found in the literature that solves gradient poisoning attacks in federated machine learning; however, we have tried to cover up most of the approaches proposed by researchers to solve this attack. The approaches proposed in the literature to solve these attacks can be categorized into two categories, i.e., filtering based approach, and blockchain-based approach. We have briefly discussed these two categories of approaches below.

In a filtering-based approach, clients validate the gradients obtained from the global model in federated training with their local dataset. The loss function score is calculated during validation, and upon the analysis of clients that have contributed negatively to the global model, adversaries are identified and then blocked. Authors at [11] have discussed an approach called FoolsGold that can solve the poisoning attacks done by the sybils. In their approach, they have calculated the cosine similarity between the updated gradients. On the basis of similarities score, they identify the gradients of sybils and non-sybils. Their approach is highly efficient even if the number of participating sybils is large and does not require further information from clients rather than updated gradients. However, their proposed approach cannot detect the poisoning attack if the number of participating sybil is one. Author at [12] proposes an approach that is based on the Krum function. Krum function does not tolerate a single byzantine failure. Authors at [11] have combined the Krum function along with the FoolsGold to solve the single sybil based poisoning attacks. Authors at [13] have proposed an approach where they identify the malicious model during federated training, and these models are eliminated before they make any contribution to global averaging. The malicious model is identified by taking the loss function into consideration by validating the global model with the client's local dataset. The clients with higher loss functions are identified as sybils.

In a blockchain-based approach, the updated gradients are first passed through the blockchain-based authentication protocol. In an adversarial environment, if the gradient filtration protocol is at client side, then adversaries can gain control over this protocol and evade detection. However, in a blockchain-based approach, the integrity of the gradients is not checked by one node; instead, it is checked by thousands of nodes that are decentralized around the world. In this approach, even if the adversary has control over one node that serves as the gradient authentication protocol, through the consensus protocol, the validity of the false gradients is ignored, and hence the poisoning of the global model is reduced. Authors at [14] have proposed a fully decentralized peer to peer blockchain-based architecture that uses blockchain and cryptographic protocols to verify the gradients. Their proposed approach is very scalable and can filter out poisoned gradients; even more than 30% of the sybills try to poison the global model at once. Authors at [15] have proposed a digital signature based verification protocol that identifies each client aiming to corrupt a global model.

## III. Background

To better understand the gradient poisoning attack and our proposed approach to solve this attack, the readers need to know how federated training and averaging works in machine learning. In this section, with the help of Algorithm 1, we have given a detailed discussion on those topics.

In our approach, we have used the federated averaging algorithm proposed by McMahan [9]. The SGD based federated averaging algorithm is described in Algorithm 1. In algorithm 1, initially, all the nodes participating in training receive some weights to their ML algorithm parameters. In the case of SGD, we have used learning rate η as a parameter. Each participating node has its local dataset, which is denoted by $d_n$. The model's parameters at the client side are then updated with the weight received from the global server, and then the model is trained on the dataset, i.e., $d_n$ of each client. The weight that is obtained after the model fits perfectly on each client dataset is denoted by $W_{weight}$. The $W_{weight}$ of each client is then sent to the global server for averaging. $W_{average}$ denotes the average weight in algorithm 2. The averaged weight is again sent to all the participating nodes for the next round of training. The process of federated averaging and training continues until a global model's loss function is converged to the desired value. An adversary can access a client and participate in a federated averaging by injecting the false gradients. Here false gradients mean any gradients whose magnitude deviates significantly from the true gradients and can largely decrease model accuracy.

**Algorithm 1.** Federated averaging SGD algorithm

**Procedure** Sever()
    $W_{all\_nodes}$ = Receive_weight()
    $W_{average} \leftarrow \frac{1}{n} \sum_{k=1}^{n} W_k$
    Set weight of $M_{global} \leftarrow W_{average}$
    Send($M_{global}$)

**Procedure** Node()
    D ← Local dataset divided into mini datasets $d_n$
    $M_{gloabl}$ = Receive(global_model)
    **for** all each node in K in parallel **do**
        Train $M_{global}$ with their respective dataset $d_n$
        $W_{weight} = W_{weight} - \eta \Delta l(W_{weight}, d_n)$

$W_{updated\_weight} \leftarrow W_{weight}$
Send_weight($W_{updated\_weight}$)

## IV. PROPOSED APPROACH

In our proposed approach, we have developed a gradient filtration protocol that filters the gradients coming from adversaries using clustering and reward based algorithms. Our proposed filtration protocol is shown in figure 1. To give a reader a clear view of our proposed approach, we have discussed in detail our architecture with the help of following sections.

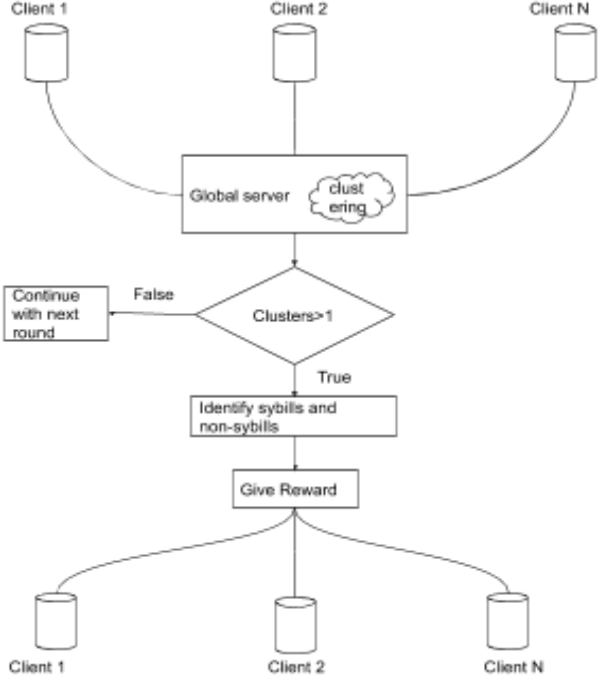

Fig 1. Proposed gradient filtering architecture

### A. Clustering of gradients

In our clustering-based approach, we form different clusters of incoming gradients from the clients participating in federated training. Creating different clusters helps in identifying the gradients which have deviated from original gradients. In federated training we can expect no adversaries at most one cluster where that cluster represents the gradients from benign clients. However, in an adversarial environment, these clusters can be more than one. It depends on the randomness in the magnitude of the gradients an adversary is injecting to the global model during federated training. Our assumption relies on the fact that there is a high chance that there are adversaries during federated training if there are more than one cluster. The result presented in section VI proves our assumption.

To cluster the gradients, we have used the K-means clustering algorithm [16]. K means clustering is an unsupervised learning algorithm where the main intuition is to find the number of meaningful clusters, i.e., K. K means algorithm works in two phases. In the first phase, the neighbouring datasets are assigned to the centroids, and in the second phase, the same centroid is relocated by calculating its new position. This process continues until there are no data points that can be reassigned to the new centroids. Let $\{(x^1, y^1),(x^2, y^2),(x^3, y^3),.......,(x^m, y^m)\}$ be unlabeled gradients such that $x^i \in R^d$, where $R^d$ represents the d dimensional vector of a dataset. Let K be the number of clusters with the centroids $\mu_1,\mu_2,....,\mu_k$. Initially, all the centroids are randomly initialized in the X-Y coordinate. All the centroid calculates the euclidean distance between its neighbouring gradients $x^i$ and centroid $\mu_k$ using equation 1. In equation 1 $c^{(i)}$ is the index of clusters (1,2,...,K).

$$c^{(i)} = \min \| x^{(i)} - \mu_k \|^2 \quad (1)$$

While calculating the distance, the main intuition is to minimize the cost function J, which is represented by equation 2.

$$J(c^{(1)},......,c^{(m)}, \mu_1,......,\mu_k) = \frac{1}{m} \sum_{i=1}^{m} \| x^{(i)} - \mu_k \|^2 \quad (2)$$

Once the gradient of $x^{(i)}$ is assigned to its closest centroid $\mu_k$, the mean of the location of the assigned gradients, i.e., $\mu_{center}$ to the centroid $\mu_k$ is calculated, which is given by $\mu_{center} = \frac{1}{m} \sum_{i=1}^{m} (x^i)$. As described in algorithm 2, we generally find only one cluster in an adversary free federated training. However, the number of clusters may increase in the presence of the adversaries. Let $D_i$ and $D_i^{'}$ be the clusters representing the gradients of the benign and adversary clients, then their dissimilarity function is given in equation 3.

$$\Delta(D_i, D_i^{'}) = \sum_{j=1}^{L} \Delta_j(D_i, D_i^{'}) \quad (3)$$

When an adversary participates in federated training, the dissimilarity value becomes very high because the adversaries inject the gradients that are very high or low in magnitude from original gradients. High dissimilarity score results in the formation of more than one cluster. Formation of clusters gives information about the adversaries node_id, ip address and geolocation. In our experimental evaluation, we have only extracted the node_id of the adversaries.

**Algorithm 2**. Clustering and rewarding algorithm

---

**Procedure** Clustering()
    normal_cluster = 1
    **for** all each **round** in training round **do:**
        gradients_all = obtain_gradients()
        **for** all each **gradient** in gradeints_all **do:**
            clusters = calculate clusters()
        **if** clusters > normal_cluster
            adversaries=get_adversaries_nodes
            benign=get_benign_nodes
        **else**
            benign=get_benign_nodes

**Procedure** Reward()
    client_score_list=[]
    **for** all each **clients** in adversaries and benign **do**:
        **if** clients is adversaries do:
            client_score_list[client]=negative
        **else**
            client_score_list[client]=positive

**Procedure** Elimination()
    **for** all each **client_score,client_id** in client_list **do**:
        **if** client_score > threshold
            eliminate_client(client_id)

*B.     Rewarding and eliminating clients*

Once the node_id of the adversaries and benign clients are identified, as described in algorithm 2, the benign clients are awarded positive scores, and the adversaries are awarded negative scores. We maintain the threshold of the negative score, after which we assume that if any clients cross the threshold score, they are the adversaries. Identifying permanent adversaries and its elimination is very important in federated training because if we do not eliminate those adversaries, clients will need several rounds for their model to converge. Elimination converges this global model faster, and the experimental result in section VI proves this theory.

V.     EXPERIMENTATION

Our experimentation was performed on a machine having a core i5 processor with a clock speed of 3.4GHz, 8 GB of RAM, and 2 GB of the graphics card. We have used python for the complete implementation of our federated machine learning model and our proposed approach. We have used the NSL-KDD dataset to build the federated machine learning-based IDS and to validate our approach. The main intuition behind building IDS was to see the impact that adversaries can produce in deviating IDS detection capability by poisoning the gradients during federated training. NSL-KDD contains 125,973 records in KDDTrain+ file [17]. We divided the records present in the file among our ten participating clients. We only choose ten participating clients; however, in real life, the number of adversaries participating in federated training is in millions of numbers. The distribution of data is in a Non-IID manner, which is presented in table I.

Table I: Distribution of dataset among 10 clients

| Dataset | Attack Type | Number of instances |
|---|---|---|
| Client_1 | neptune, smurf | 40466 |
| Client_2 | land, teardrop | 6107 |
| Client_3 | pod, back | 6315 |
| Client_4 | portsweep, nmap | 8938 |
| Client_5 | ipsweep, satan | 11135 |
| Client_6 | Imap, warezmaster | 5420 |
| Client_7 | ftp_write, guess_passwd | 5442 |
| Client_8 | multihop, spy | 5402 |
| Client_9 | phf, warezclient, buffer_overflow | 6135 |
| Client_10 | loadmodule, perl, rootkit | 5411 |

To give a reader a clear idea about the effect of gradient poisoning attack in federated training, we launched an attack by injecting the gradients which were very much dissimilar in magnitude from the original gradients. Adversaries can poison the federated training either by manually injecting the false gradients or can obtain it by adding some noise mechanism [18]. In our case, we have used the Laplace noise to add the noise in the gradients. We have used the Diffprivlib library developed by IBM to add laplacian noise [19]. Let $G_{original}$ be the original gradients; then the poisoned gradient $G_{poisoned}$ is obtained from equation 4.

$$G_{poisoned} = Lap(G_{original} \mid \mu, b) = \frac{1}{2b} \exp\left(\frac{-|x-\mu|}{b}\right) \quad (4)$$

In the above equation, µ tells us about the position of distributions that is either positive or negative, and b is an exponential scale parameter that gives an idea about the distribution of a laplacian noise [18]. We have set b=0.005 to add a high level of noise to the gradients. In our experiment, we assume that adversaries can launch attacks in two different scenarios. In the first scenario, adversaries can control the clients and make clients act as an adversary. In the second scenario, an adversary can itself participate in federated training using a large number of IoT devices. For the first scenario, we assume that 40% of the clients are adversaries; among all the participating clients and are poisoning the gradients during federated training. In the second scenario, it becomes very difficult for the adversaries to collect a large number of IoT devices and participate in federated training, so we assume that there 20% of the participating clients are adversaries. Our proposed algorithm is fetched in a global server to filter the participating adversaries, considering the low processing power of IoT devices. All the computation is done at the server-side reducing the need for computation at a client-side.

VI.     RESULTS AND DISCUSSION

Figure 3 gives an idea about the decrease of accuracy in an adversarial federated IoT environment. From figure 2, we can get an idea that the maximum accuracy up to 99.8% can be achieved in adversarial free-federated training. In contrast, figure 3 represents a significant decrease in accuracy in the presence of adversaries to as low as 70%. From figure 2 and figure 3, we can compare that the model converges faster in about 60 rounds where there are no adversaries, whereas the model is taking longer rounds, i.e., up to 80 in convergence during federated training. In the IoT environment, we aim to develop a machine learning-based approach which consumes less memory and power, but since these kinds of attacks need larger rounds to converge, it may extensively consume resources in IoT devices. In figure 3, we can see that a maximum decrease in accuracy can be seen when more adversaries participate in federated training than the less number of adversaries participating in federated training.

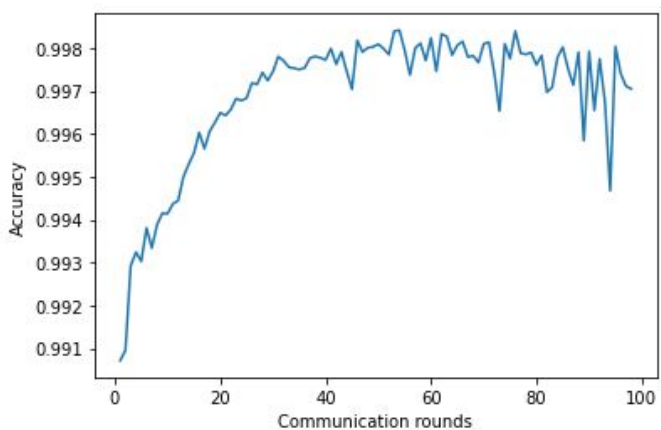

Fig 2. Accuracy in a adversarial federated environment

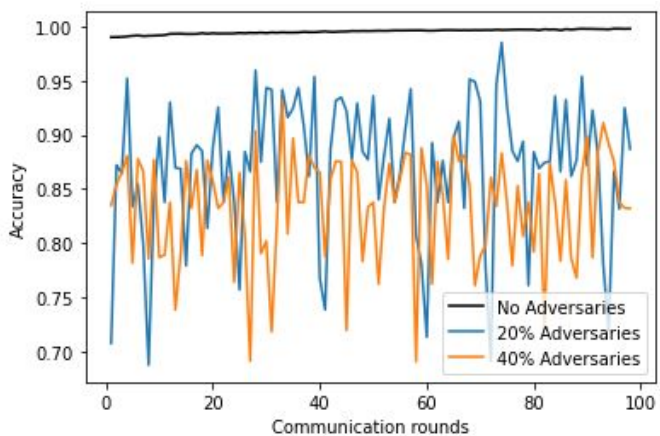

Fig 3. Accuracy in a adversarial federated environment

To detect the adversaries, we clustered the gradients obtained from participants in each round. From figure 4, we can see that when there are no adversaries, only one cluster is formed. This is because in federated training, we obtain weights from the clients with similar datasets when dealing with the same kind of machine learning problem. However, when there is the presence of adversaries, the adversaries do not have an idea about the weight of the gradient that is being exchanged between the server and the clients, and any injection of the false gradients will result in a larger number of clusters. In figure 4, the formation of 2 clusters can be seen in the presence of 20% and 40% adversaries, which clearly proves our assumptions. In figure 4, the WCSS score on Y axis is a sum of variance between the observations in each cluster. The clustering problem has an ideal number of clusters when there is no more decrease in WCSS score in an increasing number of clusters. Figure 4, shows that this score largely decreases in two clusters, mediumly in three clusters, and minimally after three clusters. Hence, in our case, two clusters are optimal number of clusters. However, there can be more than two numbers of clusters in adversarial federated training. The number of clusters depends upon the magnitude of the weight of gradients adversaries are injecting during a federated training.

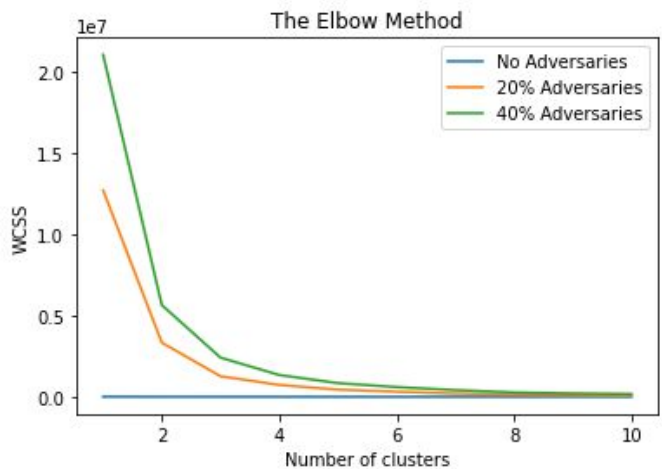

Fig 4. Number of cluster formed in a adversarial and non-adversarial federated environment

Clustering the gradients in every round can help in getting an idea about the ID of honest and malicious clients. In our experiment, according to algorithm 2, if a certain client crosses the threshold of a negative reward, they are completely eliminated from the training. We have set this threshold score to 20, and from figure 5, we can see that when this 40% of the adversaries are participating, there is a decrease in accuracy, but once they cross the threshold reward, they are completely eliminated from training. Figure 5 shows that eliminating the clients in further rounds has increased the accuracy of a global model in federated training.

Our gradient filtration protocol runs on the server-side and doesn't need any extra processing power from the client-side. This makes our protocol very suitable for IoT devices. Moreover, our protocol does not need to interact with any extra servers for gradient authentication, as described by authors in their blockchain-based gradient authentication protocol [14,15]. The minimum number of interactions between the outside servers decreases the bandwidth consumption of IoT devices.

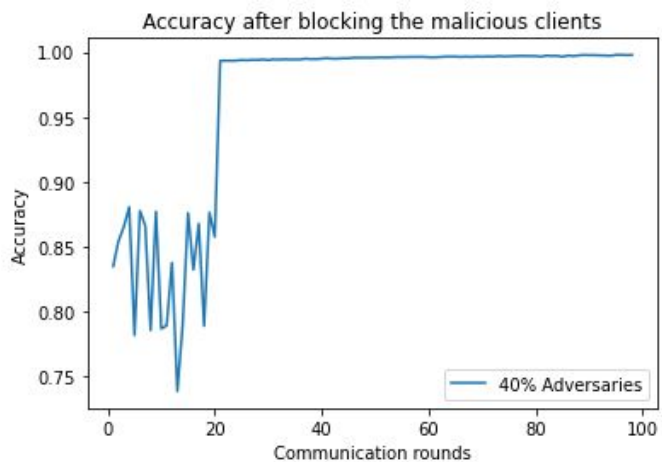

Fig 5. Accuracy after blocking the malicious clients

## VII. Conclusion

In this paper, we discussed the unsupervised clustering algorithm based approach in detecting adversaries in a federated machine learning-based IoT environment. With the help of experimental results, we concluded that the presence of adversaries can be very harmful in deteriorating global model accuracy and can utilize maximum resources present in IoT devices. Keeping in a mind the resource constraint nature of IoT devices, we developed the gradient filtration protocol which was very efficient in detecting adversaries and eliminating them. In the coming future, we would like to leverage our gradient filtration protocol with some sophisticated deep learning algorithms that can be feasible for IoT devices.